\documentclass{article}

\usepackage{arxiv}

\usepackage[utf8]{inputenc} 
\usepackage[T1]{fontenc}    
\usepackage{hyperref}       
\usepackage{url}            
\usepackage{booktabs}       
\usepackage{amsfonts}       
\usepackage{nicefrac}       
\usepackage{microtype}      
\usepackage{xcolor}         
\usepackage{microtype}
\usepackage{graphicx}
\usepackage{subfigure}
\usepackage{makecell}
\usepackage{multirow}
\usepackage{amsmath, amsthm, amssymb}
\usepackage{hyperref}
\usepackage{amsmath}
\usepackage{amssymb}
\usepackage{mathtools}
\usepackage{amsthm}
\usepackage[capitalize,noabbrev]{cleveref}
\usepackage{enumitem}
\usepackage{algorithm}
\usepackage{algorithmic}
\theoremstyle{plain}
\newtheorem{theorem}{Theorem}[section]
\newtheorem{proposition}[theorem]{Proposition}

\theoremstyle{definition}

\theoremstyle{remark}

\title{GAN-based Vertical Federated Learning for Label Protection in Binary Classification}

\author{%
Yujin Han\\
Department of Biostatistics\\
Yale University\\
New Haven, CT 06520, USA\\ \texttt{yujin.han@yale.edu} \\
\And
Leying Guan\\
Department of Biostatistics\\
Yale University\\
New Haven, CT 06520, USA\\ \texttt{leying.guan@yale.edu} \\
}

\begin{document}
\maketitle
\begin{abstract}
Split learning (splitNN) has emerged as a popular strategy for addressing the high computational costs and low modeling efficiency in Vertical Federated Learning (VFL). However, despite its popularity, vanilla splitNN lacks encryption protection, leaving it vulnerable to privacy leakage issues, especially Label Leakage from Gradients (LLG). Motivated by the LLG issue resulting from the use of labels during training, we propose the Generative Adversarial Federated Model (GAFM), a novel method designed specifically to enhance label privacy protection by integrating splitNN with Generative Adversarial Networks (GANs). GAFM leverages GANs to indirectly utilize label information by learning the label distribution rather than relying on explicit labels, thereby mitigating LLG. GAFM also employs an additional cross-entropy loss based on the noisy labels to further improve the prediction accuracy. Our ablation experiment demonstrates that the combination of GAN and the cross-entropy loss component is necessary to enable GAFM to mitigate LLG without significantly compromising the model utility. Empirical results on various datasets show that GAFM achieves a better and more robust trade-off between model utility and privacy compared to all baselines across multiple random runs. In addition, we provide experimental justification to substantiate GAFM's superiority over splitNN, demonstrating that it offers enhanced label protection through gradient perturbation relative to splitNN.
\end{abstract}

\section{Introduction}
\label{Introduction}

Federated learning trains algorithms across multiple decentralized remote devices or siloed data centers without sharing sensitive data. There are three types of federated learning depending on the data partitioning methods used: horizontal federated learning (HFL), vertical federated learning (VFL), and federated transfer learning \cite{yang2019federated}. VFL partitions the data vertically, where local participants have datasets with the same sample IDs but different features \cite{hardy2017private}. With stricter data privacy regulations like CCPA1 \cite{pardau2018california} and GDPR3 \cite{voigt2017eu}, VFL is a viable solution for enterprise-level data collaborations, as it facilitates collaborative training and privacy protection. However, VFL faces challenges in terms of high memory costs and processing time overheads, attributed to the complex cryptographic operations employed to provide strong privacy guarantees \cite{zhang2020batchcrypt}, including additive homomorphic encryption \cite{hardy2017private} and secure multi-party computation \cite{mohassel2017secureml}, which are computationally intensive. To address these issues, split learning \cite{gupta2018distributed,vepakomma2018split,abuadbba2020can} has emerged as an efficient solution, allowing multiple participants to jointly train federated models without encrypting intermediate results, thus reducing computational costs. SplitNN \cite{ceballos2020splitnn}, which applies the concept of split learning to neural networks, has been used successfully in the analysis of medical data \cite{poirot2019split,ha2021spatio}.

Split learning, while reducing computational costs, poses substantial privacy risks due to the absence of encryption protection for model privacy. 
One specific privacy risk is Label Leakage from Gradients (LLG) \cite{wainakh2022user}, in which  gradients flowing from the label party to the non-label (only data) party can expose the label information \cite{erdogan2021splitguard, zhu2019deep,wainakh2021user}. LLG is susceptible to exploitation for stealing label information in binary classification and has limited proposed solutions to address it. Recent work by Li et al. \cite{li2021label} indicates that \textit{in binary classification, the gradient norm of positive instances is generally larger than negative ones}, which could potentially enable attackers to easily infer sample labels from intermediate gradients in splitNN. Despite the fact that binary classification is widely used in various federated scenarios, such as healthcare, finance, credit risk, and smart cities \cite{crowson2022systematic,byrd2020differentially,cheng2021secureboost,zheng2022applications}, and is vulnerable to LLG, limited research has been conducted on addressing the LLG issue in binary classification. Previous studies \cite{ceballos2020splitnn,erdogan2021splitguard,titcombe2021practical,pereteanu2022split} have focused mainly on securing the data information of non-label parties, while ignoring the risk of leaking highly sensitive label information of the label party. Therefore, it is critical to address how splitNN can resist LLG in binary classification tasks. 

In this work, in order to prevent inferring sample labels from the gradient calculation, we introduce a novel Generative Adversarial Federated Model (GAFM), which synergistically combines the vanilla splitNN architecture with Generative Adversarial Networks (GANs) to indirectly incorporate labels into the model training process. Specifically, the GAN discriminator within GAFM allows federated participants to learn a prediction distribution that closely aligns with the label distribution, effectively circumventing the direct use of labels inherent in the vanilla SplitNN approach. Moreover, to counteract the potential degradation of model utility induced by GANs, we enhance our method by incorporating additional label information via an additional cross-entropy loss, which encourages the intermediate results generated by the non-label party and the labels with added noise provided by the label party to perform similarly. The entire framework of the proposed GAFM and the training procedures is displayed in Figure \ref{fig:GAFM Pipeline}. Our contributions are highlighted as follows.

\begin{itemize}[leftmargin=*]
    \item We propose a novel GAN-based approach, called GAFM, which combines vanilla splitNN with GAN to mitigate LLG in binary classification (section \ref{Generative Adversarial Federated Model(GAFM)}). Our analysis in section \ref{Can GAFM protects label stealing attacks?} demonstrates that GAFM protects label privacy by generating more mixed intermediate gradients through the mutual gradient perturbation of both the GAN and cross-entropy components.
    
    \item We enrich the existing gradient-based label stealing attacks by identifying two additional simple yet practical attack methods, namely mean attack and median attack in section \ref{Label stealing Attacks}. The experimental results in section \ref{Evaluation of utility and privacy} demonstrate that our new attacks are more effective in inferring labels than the existing ones.
    
    \item We evaluate the effectiveness of GAFM on various datasets. Empirical results in section \ref{experiments} show that GAFM mitigates LLG without significant model utility degradation and the performance of GAFM across different random seeds is more stable compared to baselines. We also provide additional insights based on the ablation experiment (section \ref{Ablation study}) to demonstrate the necessity of combining both the GAN and cross-entropy components in GAFM. Compared to the stable balance between utility and privacy that can be achieved by using both components, GAFM with only the GAN component provides enhanced privacy protection at the cost of reduced utility, while GAFM with only the cross-entropy component delivers superior utility but offers limited privacy protection.

\end{itemize}

\section{Related Work}
\label{Related Work}
\noindent\textbf{SplitNN-driven Vertical Partitioning.} SplitNN enables multiple participants to train a distributed model without sharing their data and encrypting intermediate results \cite{gupta2018distributed, vepakomma2018split, ceballos2020splitnn}. In SplitNN, each passive participant trains a partial neural network locally, and the layer at which the label and non-label participants share information is called the cut layer. At the cut layer of SplitNN, each non-label party trains a fixed portion of the neural network locally and shares intermediate results with the label party. The label party then aggregates these intermediate results and implements backward propagation to update the local parameters of each non-label party. There are several methods of aggregation, such as element-wise average, element-wise maximum, element-wise sum, element-wise multiplication, concatenation, and non-linear transformation \cite{ceballos2020splitnn}.

\noindent\textbf{Label privacy protection via random gradient perturbation.} To address the issue of LLG through intermediate gradients at the cut layer, one possible solution is to introduce randomness to the intermediate gradients, which has been utilized in HFL \cite{abadi2016deep,geyer2017differentially,hu2020personalized}. \textbf{Marvell} \cite{li2021label} is a random perturbation approach specifically designed for binary classification tasks. Marvell protects label privacy by perturbing the intermediate gradients and aims to find the optimal zero-centered Gaussian perturbations that minimize the sum of KL divergences between two perturbed distributions, while adhering to a budget constraint on the amount of perturbation added: 
\begin{equation}
\label{marvell definition}
\begin{split}
 &\min_{W^{(0)}, W^{(1)}} \rm{KL}(\tilde{\mathbb P}^1\| \tilde{\mathbb P}^0)+\rm{KL}(\tilde{\mathbb P}^0\| \tilde{\mathbb P}^1)\quad \\ &\rm{s.t.}\quad p \rm{tr}(\boldsymbol{\Sigma}_0)+(1-p)\rm{tr}(\boldsymbol{\Sigma}_1) \leq P,
\end{split}
\end{equation}
where $\tilde{\mathbb P}^k$ is the distribution of perturbed intermediate gradients from $k$ after convolution with the Gaussian noise $W^{(k)}=\mathcal{N}(0, \boldsymbol{\Sigma}_k)$, $k=0,1$, p is the weight and P is the budget for how much random perturbation Marvell is allowed to add. As the intermediate gradients with different labels become less distinguishable, it becomes more difficult for attackers to infer labels from gradients.

\textbf{Max Norm} \cite{li2021label} is an improved heuristic approach of adding zero-mean Gaussian noise with non-isotropic and example-dependent covariance. More concretely, for the intermediate gradient $\mathbf g_j$ of data point $j$, Max Norm adds the zero-mean Gaussian noise $\eta_j$ to it with the covariance as:
\begin{equation}
\label{maxnorm definition}
\begin{split}
\sigma_j=\sqrt{\frac{\|\mathbf g_{\mathrm{max}}\|^2_2}{\|\mathbf g_j\|^2_2}-1},
\end{split}
\end{equation}
where $\|\mathbf g_{\mathrm{max}}\|^2_2$ is the largest squared 2-norm in a batch. Max Norm is a simple, straightforward, and parameter-free perturbation method. But it does not have strong theoretical motivation and cannot guarantee to defend unknown attacks \cite{li2021label}.


\textbf{GAN in Federated Learning.} Previously, researchers have combined GAN and HFL for various purposes, including three directions in HFL: (1) generating malicious attacks \cite{hitaj2017deep,wang2019beyond}, (2) training high-quality GAN across distributed data under privacy constraints \cite{hardy2019md,rasouli2020fedgan,mugunthan2021bias}, and (3) protecting client data privacy \cite{wu2021fedcg}, known as FedCG. In FedCG, each client train a classifier predicting the response $Y$ using  $Z$ extracted from the original feature $X$, and a conditional GAN that learns the conditional distribution of $Z$ given $Y$. The classifiers and generators (instead of the extracted $Z$) from different clients are then passed to the server for updating the common model. FedCG has been shown to effectively protect clients' data privacy in HFL. These previous works have demonstrated the promising role of GAN in HFL, but to the best of our knowledge, there has been no prior work investigating the GAN model for label protection in VFL.

\section{Generative Adversarial Federated Model (GAFM)}
\label{method}

In this section, we first introduce the split learning problem for binary classification, including its associated notations. We then provide a detailed description of the GAFM method, along with a discussion on GAFM's parameter selection. Additionally, we explain how GAFM offers better label privacy protection than vanilla splitNN and propose new label stealing methods to evaluate GAFM's privacy protection ability in subsequent experiments.

\begin{figure}
\centering
\includegraphics[width=14cm,height=7.5cm]{
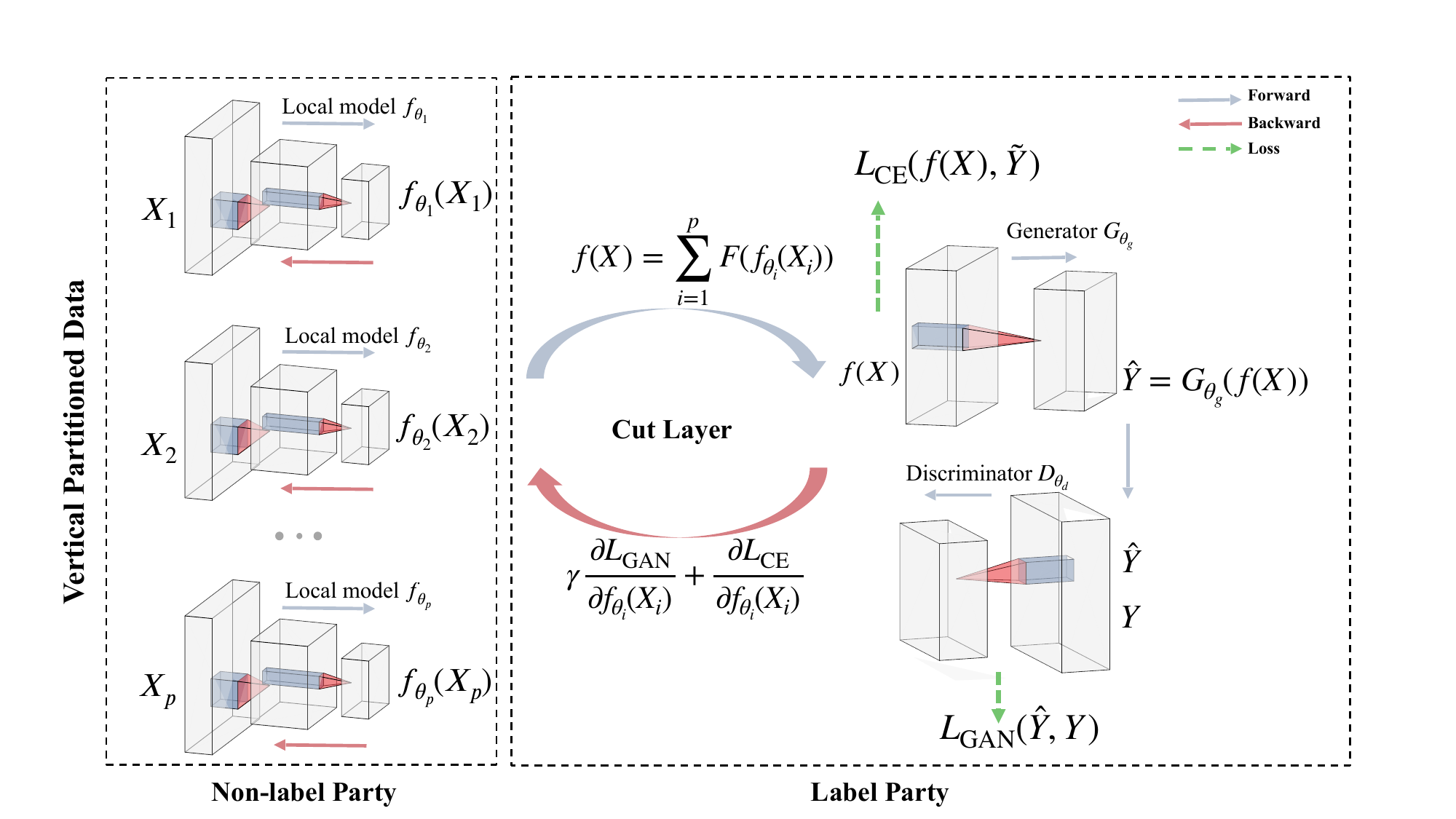}
\caption{An illustration of \textbf{Generative Adversarial Federated Model (GAFM)} with $P$ non-label parties and one label party. Each non-label participant uses its local model $f_{\theta_p}$ to extract feature information $f_{\theta_p}(X_p)$ from its local data $X_p$, where $\theta_p$ is denoted as the local model parameter of the $pth$ non-label participant. The label party aggregates all feature information in the cut layer to obtain intermediate results $f(X)$, which are then used as input to the generator $G_{\theta_g}$ with $\theta_g$ being the generator model parameter. The label party then trains the generator $G_{\theta_g}$ and discriminator $D_{\theta_d}$ adversarially using the true label $Y$ and the prediction $\hat Y$ (the output of $G_{\theta_g}$), where $\theta_d$ is the discriminator model parameter. To incorporate additional true label information for improving the prediction accuracy, an additional cross-entropy loss $L_{\mathrm{CE}}$ is introduced. Section \ref{method} provides more details on GAFM.}
\label{fig:GAFM Pipeline}
\end{figure}

\subsection{Problem setting and notation}
\label{Problem setting and notation}

We consider the joint training of one label party and $P$ non-label parties for binary classification tasks over the domain $\mathcal{X} \times \{0,1\}$. Each non-label party $p$ possesses one local model $f_{\theta_p}: \mathcal{X} \to \mathbb{R}^d$ where $\theta_p$ is the local model parameter of non-label party $p$ and the local data $X_p \in \mathcal{X}$. In vanilla splitNN, the label party owns the transformation function $F: \mathbb{R}^d \to \mathbb{R}^d$ for feature aggregation in the cut layer, the logit function $h: \mathbb{R}^d \to \mathbb{R}$ for prediction, and each example's label $y \in \{0,1\}$. In contrast, in GAFM, the transformation function $F$ for feature aggregation remains with the label party, while the logit function $h$ is replaced by the generator $G_{\theta_g}: \mathbb{R}^d \to \mathbb{R}$ where $\theta_g$ is the generator model parameter. Furthermore, GAFM includes an additional discriminator $D_{\theta_d}:\mathbb{R} \to \mathbb{R}$ where $\theta_d$ is the discriminator model parameter.

\subsection{GAFM framework}
\label{Generative Adversarial Federated Model(GAFM)}

In this section, we provide a detailed description of the GAFM framework. As illustrated in Figure \ref{fig:GAFM Pipeline}, each non-label party $p$ in GAFM utilizes its local model $f_{\theta_p}$ to extract local feature information from its local data $X_p$ and sends it to the label party. In the cut layer, the label party aggregates the feature information from all non-label parties to obtain the intermediate result $f(X) = F(\sum_p f_{\theta_p}(X_p))$, where $F(.)$ is a known transformation function, such as the identity function. The label party takes the intermediate result $f(X)$ as input to the generator $G_{\theta_g}$, and the output of $G_{\theta_g}$ is treated as the prediction $\hat{Y}$. Then, the label party adversarially trains both the generator $G_{\theta_g}$ and the discriminator $D_{\theta_d}$ with guidance from the true label $Y$ and the prediction $\hat{Y}$. The GAN loss $L_{\mathrm{GAN}}$ between $G_{\theta_g}$ and $D_{\theta_d}$ is defined in equation \ref{eq:minmax_GAN_obj}, which quantifies the distance between the prediction $\hat Y$ and the ground truth $Y$ distribution by employing the Wasserstein-1 distance \cite{kantorovich1960mathematical}:
\begin{equation}
\label{eq:minmax_GAN_obj}
\begin{split}
L_{\mathrm{GAN}}(\theta_d,\theta_g;\varepsilon)&=\mathbb E_{Y}\left[D_{\theta_d}(Y+\varepsilon)\right]-\mathbb E_{X}\left[D_{\theta_d}(\hat{Y})\right],
\end{split}
\end{equation}
where $\varepsilon\sim N(0, \sigma^2)$ is a small  additive Gaussian noise so that $Y+\varepsilon$ has a continuous support. The Wasserstein loss, which encourages the predicted distribution to closely align with the empirical response, is applicable to classification tasks \cite{frogner2015learning}.

To improve the prediction accuracy in GAFM, we introduce an additional cross-entropy (CE) loss, which measures the distance between the intermediate result $f(X)$ and the randomized response $\tilde{Y}$. The randomized response $\tilde{Y}$ is defined as: 
\begin{equation}
\label{eq:randomized response}
\begin{split}
\tilde{Y}=\begin{cases} 0.5+u;Y=1\\
0.5-u;Y=0, \end{cases}
\end{split}
\end{equation}
where the random variable $u \sim \mathrm{Uniform}(0,\Delta)$ and the hyperparameter $\Delta \in [0,0.5]$. The randomized response $\tilde Y$ replaces the original response $Y$, which enhances label protection against LLG. Setting $\Delta=0$ removes any label information from the original data, while $\Delta>0$ allows $\tilde{Y}$ to guide the classifier without excessive label leakage from CE loss. The use of a randomized response is a departure from previous work, which used the true label $Y$ and encountered issues with label leakage. Following the equation \ref{eq:randomized response}, the CE loss is:
\begin{equation}
\label{eq:minmax_recon_obj}
\begin{split}
L_{\mathrm{CE}}(\boldsymbol{\Theta};u)&=-\mathbb E_{XY}[\tilde{Y}\log(\sigma(f(X)))
+(1-\tilde{Y})\log(1-\sigma(f(X))].
\end{split}
\end{equation}
where $\boldsymbol{\Theta}=\{\theta_1,\dots,\theta_P\}$ is the local model parameter set and the function $\sigma$ is the sigmoid function. In particular, $\tilde Y \in \mathbb{R}$ and $f(X)\in \mathbb{R}^d$, where $d$ is an arbitrary dimension. When $d \neq 1$, $f(X)$ needs to be summed along the $d$-dimension to be consistent with the dimension of $\tilde Y$.

Combine the $L_{\mathrm{GAN}}$ defined in equation \ref{eq:minmax_GAN_obj} and the $L_{\mathrm{CE}}$ defined in equation \ref{eq:minmax_recon_obj}, we obtain the full loss of GAFM, as given by equation (\ref{eq:minmax_full_obj}): 
\begin{equation}
\label{eq:minmax_full_obj}
\begin{split}
\min\limits_{\theta_g,\theta_p}\max\limits_{\theta_d}L_{\mathrm{GAFM}}(\theta_d,\theta_g,\boldsymbol{\Theta};\varepsilon, u, \gamma)&=\gamma L_{\mathrm{GAN}}(\theta_d,\theta_g;\varepsilon)+L_{\mathrm{CE}}(\boldsymbol{\Theta};u),
\end{split}
\end{equation}
where the loss weight hyperparameter $\gamma >0$. 

Following the parameter update sequence of Wasserstein-based GANs \cite{arjovsky2017wasserstein}, GAFM enables the label party to update the discriminator parameter $\theta_d$ first, followed by updating the generator parameter $\theta_g$. Then, the non-label parties update local model parameters $\boldsymbol{\Theta}$:

\noindent (1) Update $\theta_d$ to maximize the GAN loss: $\theta_d\leftarrow \arg\max_{\theta_d} L_{\mathrm{GAN}}(\theta_d, \theta_g;\varepsilon)$, given $\theta_g$, $\boldsymbol{\Theta}$.

\noindent (2) Update $\theta_g$ to minimize the GAN loss: $\theta_g\leftarrow \arg\min_{\theta_g} L_{\mathrm{GAN}}(\theta_d, \theta_g;\varepsilon)$, given $\theta_d$, $\boldsymbol{\Theta}$.

\noindent (3) Update $\boldsymbol{\Theta}$ to minimize GAFM loss: $\boldsymbol{\Theta}\leftarrow \arg\min_{\boldsymbol{\Theta}} L_{\mathrm{GAFM}}( \theta_d, \theta_g,\boldsymbol{\Theta};\varepsilon,u)$, given $\theta_d$, $\theta_g$.

We employ the Adam optimizer with learning rates $\alpha_d$, $\alpha_g$, and $\alpha_p$, respectively, to implement the aforementioned procedures. To satisfy the Lipschitz constraint of the discriminator in Wasserstein-based GANs \cite{pmlr-v70-arjovsky17a, gulrajani2017improved}, we apply weight clipping to discriminator parameters $\theta_d$. This operation restricts the parameters $\theta_d$ to the interval $[-c,c]$, where $c$ is a hyperparameter. Further details regarding the algorithm can be found in Algorithm \ref{alg:GAFM}.

\subsection{Selection of GAFM hyperparameters}
\label{How to determine the hyper parameter gamma}

GAFM incorporates three crucial parameters: $\sigma$, $\gamma$, and $\Delta$. The parameter $\sigma$ introduces Gaussian noise $\varepsilon\sim N(0,\sigma^2)$ into the ground truth $Y$ distribution to render it continuously supported. The default value of $\sigma$ can be set to 0.01, and the Appendix \ref{Discussion on sigma} shows that GAFM is insensitive to different values of $\sigma$. Specifically, a broad range of $\sigma\in [0.01, 1]$ yields comparable empirical results.

The tuning of the loss weight hyperparameter $\gamma$ in GAFM can be challenging due to the vanishing gradient issue of the unnormalized GAN loss gradients \cite{pmlr-v70-arjovsky17a, gulrajani2017improved}. To address this issue, we introduce normalization techniques for the gradients of the GAN loss and the CE loss, respectively. This normalization technique in GAFM alleviates the vanishing gradient problem, making it feasible to determine the hyperparameter $\gamma$. Appendix \ref{Discussion on gamma} provides insights on selecting the appropriate value of $\gamma$ for different datasets. Our experiments demonstrate that an effective approach to determine $\gamma$ is to adjust it to balance the magnitude between the average value of the GAN loss gradient and the average value of the CE loss gradient.  Additional details on the selection of $\Delta$ are discussed in Appendix \ref{Discussion on gamma}.

We also propose a simple yet effective approach for determining the randomized response-related hyperparameter $\Delta$. Specifically, we train GAFM with different values of $\Delta$ on a 10\% subset of the data and select the optimal parameter based on its utility and privacy. The analysis in Appendix \ref{Discussion on delta}  shows that the $\Delta$ selected based on the subset is very close to or even equal to the optimal parameter selected on the full dataset. Additional details on the selection of $\Delta$ are discussed in Appendix \ref{Discussion on delta}.


\subsection{Why does GAFM protect against LLG?}
\label{Can GAFM protects label stealing attacks?}
In this section, we show that GAFM provides stronger privacy protection against LLG by generating more mixed intermediate gradients compared to vanilla splitNN. Specifically, the proposition \ref{theorem:KL} from the work \cite{li2021label} states that the KL divergence between the intermediate gradients of two classes sets an upper bound on the amount of label information an attacker can obtain via any LLG attacks.

\begin{proposition} \cite{li2021label}
\label{theorem:KL}
Let $\mathbb{\tilde P}^1$ and $\mathbb{\tilde P}^0$ be perturbed distributions for intermediate gradients of classes 1 and 0, and be continuous with respect to each other. For $\epsilon \in [0,4)$, 
\begin{equation}
\label{marvell definition}
\begin{split}
 &\rm{KL}(\mathbb{\tilde P}^1\| \mathbb{\tilde P}^0)+\rm{KL}(\mathbb{\tilde P}^0\| \mathbb{\tilde P}^1)\leq \epsilon  \\ &\mbox{implies} \quad \max_r\rm{AUC}_r\leq \frac{1}{2}+\frac{\sqrt{\epsilon}}{2}-\frac{\epsilon}{8},
\end{split}
\end{equation}
where $r$ is any LLG attack, and $\rm{AUC}_r$ represents the achieved AUC using $r$.
\end{proposition}

\begin{figure}
\centering
\centering
\includegraphics[width=14cm,height=3.6cm]{
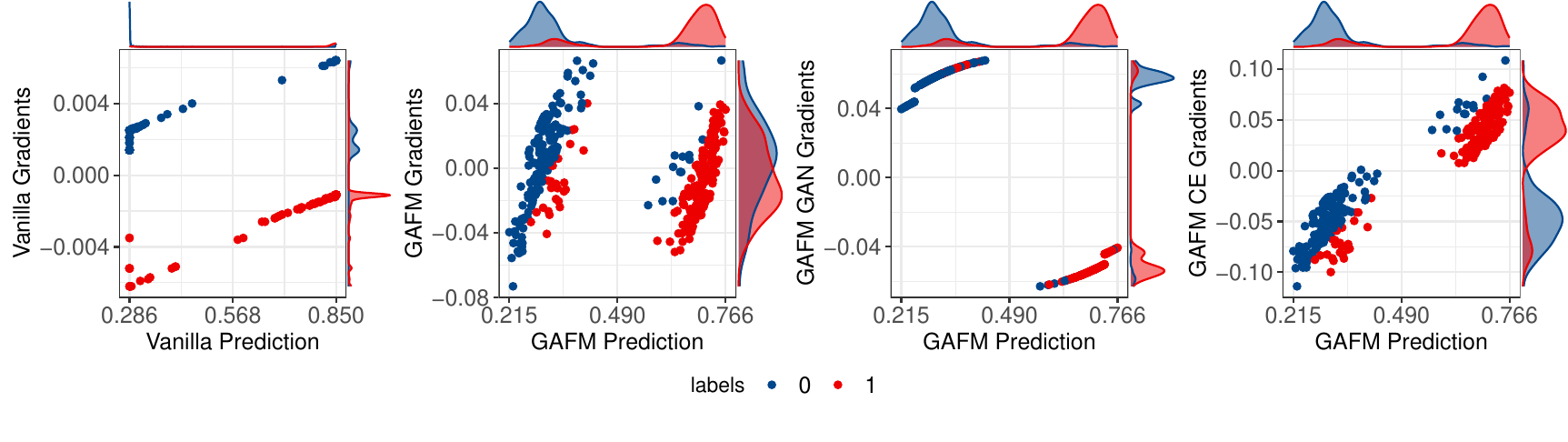}
\caption{\textbf{Comparison of Prediction vs. Intermediate Gradients } between vanilla splitNN and GAFM on IMDB. The figure \ref{fig:grads on IMDB} displays, from left to right, the intermediate gradients of vanilla splitNN, GAFM, the GAN loss gradient of GAFM, and the CE loss gradient of GAFM. Our observations show that (1) GAFM has more mixed intermediate gradients compared to vanilla splitNN; (2) the gradient class centers from the GAN loss gradient and the CE loss gradient differ in opposite directions, leading to mutual perturbation in the final gradient for GAFM. Observations from Spambase, Criteo, and ISIC in Appendix \ref{Exploring Intermediate Gradients on Additional Datasets} are consistent with those from the IMDB dataset.}
\label{fig:grads on IMDB}
\end{figure}

The proposition \ref{theorem:KL} proposes that models with more mixed intermediate gradients (meaning a smaller sum of KL divergences from the two classes) may imply a smaller upper bound $\epsilon$, which results in a smaller amount of available label information and better protection against LLG. Therefore, we indirectly demonstrate GAFM's superior label privacy protection ability by showing that GAFM has more mixed intermediate gradients than vanilla splitNN. Figure \ref{fig:grads on IMDB} illustrates the intermediate gradients of vanilla splitNN, GAFM, the GAN loss gradient of GAFM, and the CE loss gradient of GAFM on the IMDB dataset from left to right. We observe that GAFM has more mixed intermediate gradients compared to vanilla splitNN. Furthermore, the gradient class centers of the GAN loss gradient and the CE loss gradient differ in opposite directions, leading to mutual cancelation in the final gradient for GAFM. We further provide a heuristic justification in Appendix \ref{Heuristic Justification for Improved Gradients Mixing} to explain the observed discrepancy in the directions of the GAN loss gradient class centers and the CE loss gradient class centers shown in Figure \ref{fig:grads on IMDB}.

\subsection{Label stealing attacks}
\label{Label stealing Attacks}

In this section, we introduce three gradient-based label stealing attacks which are applied in subsequent experiments to evaluate the effectiveness of GAFM in mitigating LLG.

\noindent\textbf{Norm Attack.} The norm attack \cite{li2021label} is a simple heuristic for black-box attacks that can be used for label inference in binary classification tasks. The attack is based on the observation that the gradient norm $\|\mathbf g\|_2$ of positive instances tends to differ from that of negative instances, especially with unbalanced datasets. Thus, the gradient norm $\|\mathbf g\|_2$ can be a strong predictor of labels.

Previous research indicates that vanilla SplitNN gradients tend to form two clusters based on the cosine similarity sign, leading to the development of cosine attacks \cite{li2021label}. In the case of GAFM, although the relationship between the sign and the cluster of gradients may not be as straightforward, attackers can still attempt to make attacks based on proximity to a cluster. To achieve this goal, we propose mean-based attacks and further enhance robustness against outliers in gradients by introducing median-based attacks.

\noindent\textbf{Mean Attack.} We propose a mean-based attack as a heuristic for black-box attacks that exploits clustering structures in gradients. Assuming that attackers know the gradient centers of class 0 and class 1, denoted as $\boldsymbol{\mu}_0$ and $\boldsymbol{\mu}_1$, mean-based attacks assign a sample $i$ to the cluster that its intermediate gradient $\mathbf g_j$ is closer to:
\begin{equation}
\label{eq:mean-attack}
y_i=\begin{cases} 1, \;\mbox{if} \;{\|\mathbf g_i-\boldsymbol{\mu}_1\|_2} \leq {\|\mathbf g_i-\boldsymbol{\mu}_0\|_2}\\
0,\; \mbox{otherwise}. \end{cases}
\end{equation}
\noindent\textbf{Median Attack.} We propose a median-based attack, which is similar to the mean-based attack but more robust to outliers. Assuming that attackers know the gradient medians of class 0 and class 1, denoted as $ \mathbf m_0$ and $\mathbf m_1$, median-based attacks assign a sample $i$ to the cluster that its intermediate gradient $\mathbf g_j$ is closer to:
\begin{equation}
\label{eq:mean-attack}
y_i=\begin{cases} 1, \;\mbox{if} \;{\|\mathbf g_i-\mathbf m_1\|_2} \leq {\|\mathbf g_i-\mathbf m_0\|_2}\\
0,\; \mbox{otherwise}. \end{cases}
\end{equation}

\section{Experiments}
\label{experiments}

In this section, we first provide a comprehensive description of the experimental setup, including the datasets, model architectures, baselines, and evaluation metrics. Then, we present the experimental results that showcase the utility and privacy performance of GAFM . Furthermore, we conduct an ablation study to elucidate the contribution of each component of GAFM.

\subsection{Experiment Setup}
\label{Experiment Setup}

\textbf{Datasets.} Our approach is evaluated on four real-world binary classification datasets: Spambase \footnote{\url{https://archive.ics.uci.edu/ml/datasets/spambase}}, which is used for spam email discrimination; IMDB\footnote{\url{https://www.kaggle.com/datasets/uciml/default-of-credit-card-clients-dataset}}, a binary sentiment classification dataset consisting of 50,000 highly polar movie reviews; Criteo \footnote{\url{https://www.kaggle.com/c/criteo-display-ad-challenge}}, an online advertising prediction dataset with millions of examples; and ISIC \footnote{\url{https://www.kaggle.com/datasets/nodoubttome/skin-cancer9-classesisic}}, a healthcare image dataset for skin cancer prediction. Criteo and ISIC are two datasets with severely imbalanced label distributions, which have been used in the works of Max Norm and Marvell \cite{li2021label}. Furthermore, we consider another two  class-balanced datasets, Spambase and IMDB, to evaluate the performance of GAFM under different class setting. Additional information regarding the datasets and data preprocessing can be found in Table \ref{table:data_model} and Appendix \ref{dataset Processing}.

\textbf{Model Architecture.} We consider the two-party split learning setting used by Marvell, where the label party has access only to the label and the non-label party has access only to the data. In Section \ref{Multiple Clients}, we also present the utility and privacy protection results of all methods under a multi-client scenario with three non-label parties. Table \ref{table:data_model} lists the model architectures of local models $f_{\boldsymbol{\Theta}}$, generator $G_{\theta_g}$, and discriminator $D_{\theta_d}$ used for each dataset. Similar to Marvel, we employ a  Wide and Deep model \cite{cheng2016wide} for Criteo and a 6-layer CNN for ISIC as the local models $f_{\boldsymbol{\Theta}}$. More details on model architectures and training can be found in Appendices \ref{Model Architecture Details} and \ref{Model Training Details}.
\begin{table}[]
\centering
\caption{\textbf{Dataset statistics and model architectures.} Different local model architectures $f_{\boldsymbol{\Theta}}$ are considered for different datasets. 2-layer DNNs are sufficient as the generator $G_{\theta_g}$ and discriminator $D_{\theta_d}$ since their inputs are simple linear embeddings.}
\label{table:data_model}
\begin{tabular}{cccccc}
\hline
Dataset & \makecell{Positive Instance\\Proportion} &  $f_{\boldsymbol{\Theta}}$ & $G_{\theta_g}$&  $D_{\theta_d}$ \\ \hline
Spambase& 39.90\% & 2-layer DNN & \multirow{4}*{\makecell{2-layer\\ DNN}}   &  \multirow{4}*{\makecell{2-layer\\ DNN}} \\
IMDB&   50.00\% &  3-layer DNN  &    &     \\ 
Criteo&  22.66\%  &  WDL Model \cite{cheng2016wide}  &    & \\
ISIC&  1.76\% &  6-layer CNN  &    &    
\\ \hline
\end{tabular}
\end{table}

\textbf{Baselines.} We compare the utility and privacy of GAFM with Marvell \cite{li2021label}, Max Norm \cite{li2021label} and vanilla splitNN. 

\textbf{Evaluation Metrics.} We employ the Area Under Curve (AUC) metric to evaluate model utility and the leak AUC to assess model privacy protection. Differential privacy is not considered as a leakage measure, as it is not applicable to example-specific and example-aware settings like VFL \cite{li2021label}. The leak AUC \cite{li2021label, yang2022differentially, sun2022label} is defined as the AUC achieved by using specific attacks. A high leak AUC value, closer to 1, indicates that the attacker can accurately recover labels, while a low leak AUC value, around 0.5, suggests that the attacker has less information for inferring labels. To prevent simple label flipping from resulting in a higher leak AUC, we modify the leak AUC as shown in equation \ref{eq:leack-auc-0.5} for a predefined attack by flipping the label assignment if doing so results in a higher AUC:
\begin{equation}
\label{eq:leack-auc-0.5}
{\rm leak AUC} \leftarrow \max ({\rm leak AUC} , 1-{\rm leak AUC} )
\end{equation}
\subsection{Results}
\label{Results}
This section presents AUC and leak AUC to demonstrate the trade-off between privacy and utility of GAFM. We also report the results of the ablation study, which illustrates the contribution of each component of GAFM. Each method is executed 10 times with unique random seeds and train-test splits. Further experiment details can be found in the Appendix \ref{Data setup and Experiment Details}.

\subsubsection{Evaluation of utility and privacy}
\label{Evaluation of utility and privacy}
Table \ref{table:auc-leak AUC-tvd} illustrates the average AUC and average leak AUC across four datasets. GAFM achieves comparable classification AUC with Vanilla and Max Norm, but demonstrates lower leak AUC, indicating its effectiveness in protecting label privacy and defending against LLG. Compared to Marvell, GAFM achieves slightly higher average AUC on most datasets while maintaining comparable leak AUC. Importantly, GAFM demonstrates more stable performance than Marvell, with reduced variance between the worst and best AUC across different random seeds. This could be attributed to Marvell's gradient perturbation technique, which introduces Gaussian noise to intermediate gradients and can lead to unstable performance, as evidenced by the highly fluctuating leak AUC of Criteo and ISIC shown in the Marvell paper \cite{li2021label}. Additionally, we observe that our proposed mean attack and median attack achieve higher leak AUC on all four datasets compared to the norm attack. This indicates that mean attack and median attack are more effective gradient-based attacks for label stealing compared to the existing norm attack.

In conclusion, Table \ref{table:auc-leak AUC-tvd} demonstrates GAFM's improvement in privacy protection over Vanilla and Max Norm, as well as its improvement in utility and stability compared to Marvell. GAFM achieves better trade-off between utility and privacy compared to the baseline models.
\begin{table}[t]
\centering
\caption{\textbf{Comparison of utility and privacy between GAFM and baselines}. Compared to Vanilla and Max Norm, GAFM achieves lower leak AUC at similar AUC, demonstrating its effectiveness in protecting label privacy. Compared to Marvell, GAFM demonstrates comparable and even better utility and privacy protection on most datasets, while also exhibiting greater stability with reduced variance between the best and worst AUC across different random seeds. Table \ref{table:auc-leak AUC-tvd} illustrates that GAFM offers a better and more stable trade-off between utility and privacy compared to all baselines.}
\label{table:auc-leak AUC-tvd}
\begin{tabular}{lccccccc}
\hline
\multirow{2}{*}{Dataset} & \multirow{2}{*}{Method} & \multicolumn{3}{c}{Utility}        & \multicolumn{3}{c}{Privacy}                \\ \cmidrule(lr){3-5} \cmidrule(lr){6-8}  & & \makecell{Avg.\\ AUC} &\makecell{Worst\\ AUC}& \makecell{Best\\AUC} & \makecell{Norm\\ Attack} & \makecell{Mean\\ Attack} & \makecell{Median\\ Attack}\\ 
\midrule
\multirow{4}*{Spambase}&GAFM&0.93&0.91&0.95&{0.56$\pm$0.04}&{0.67$\pm$0.05}& {0.66$\pm$0.05}\\
&Marvell &0.71&0.59&0.82&{0.53$\pm$0.02}& 0.70$\pm$0.01& 0.70$\pm$0.01\\ 
&Max Norm&{0.95}&0.95&0.95&0.83$\pm$0.11&1.00$\pm$0.00&0.91$\pm$0.00\\
&Vanilla&{0.95}&0.95&0.96&0.85$\pm$0.07& 1.00$\pm$0.00&0.91$\pm$0.00\\
\midrule
\multirow{4}*{{IMDB}}&GAFM& 0.88&0.88&0.89&{0.52$\pm$0.01}&{0.60$\pm$0.01}& {0.60$\pm$0.01}\\
&Marvell&0.80&0.71&0.89& {0.52$\pm$0.01}& {0.73$\pm$0.01}& {0.73$\pm$0.01}\\ 
&{Max Norm}&{0.89}&0.89&0.90& {0.82$\pm$0.09}& 1.00$\pm$0.00& 0.99$\pm$0.01\\
&{Vanilla}&{0.89}&0.89&0.90& {0.82$\pm$0.09}& 1.00$\pm$0.00& 0.99$\pm$0.01\\
\midrule
\multirow{4}*{{Criteo}}&GAFM& 0.67&0.64 &0.73&{0.68$\pm$0.06}&{0.80$\pm$0.09}& {0.77$\pm$0.05}\\
&Marvell&0.70&0.65&0.76& {0.76$\pm$0.08}& {0.86$\pm$0.08}& {0.78$\pm$0.04}\\ 
&Max Norm&{0.69}&0.65&0.72& {0.92$\pm$0.01}& 0.83$\pm$0.09& 0.82$\pm$0.00\\
&Vanilla&{0.72}&0.69&0.77& {0.92$\pm$0.06}& 0.91$\pm$0.11& 0.82$\pm$0.00\\
\midrule
\multirow{4}*{ISIC}&GAFM& 0.68& 0.66&0.69&0.62$\pm$0.09&{0.66$\pm$0.15}&{0.68$\pm$0.11}\\
&Marvell&0.64&0.51&0.69& {0.65$\pm$0.04}& {0.69$\pm$0.01}& {0.66$\pm$0.01}\\ 
&Max Norm&0.76&0.72&0.82& {0.99$\pm$0.01}& 1.00$\pm$0.00& 0.77$\pm$0.00\\
&Vanilla&0.77&0.73& 0.82& {0.99$\pm$0.01}& 1.00$\pm$0.00& 0.78$\pm$0.00\\
\bottomrule
\end{tabular}
\end{table}

\subsubsection{Ablation study}
\label{Ablation study}

\begin{figure}
\centering
\centering
\includegraphics[width=13.6cm,height=7.3cm]{
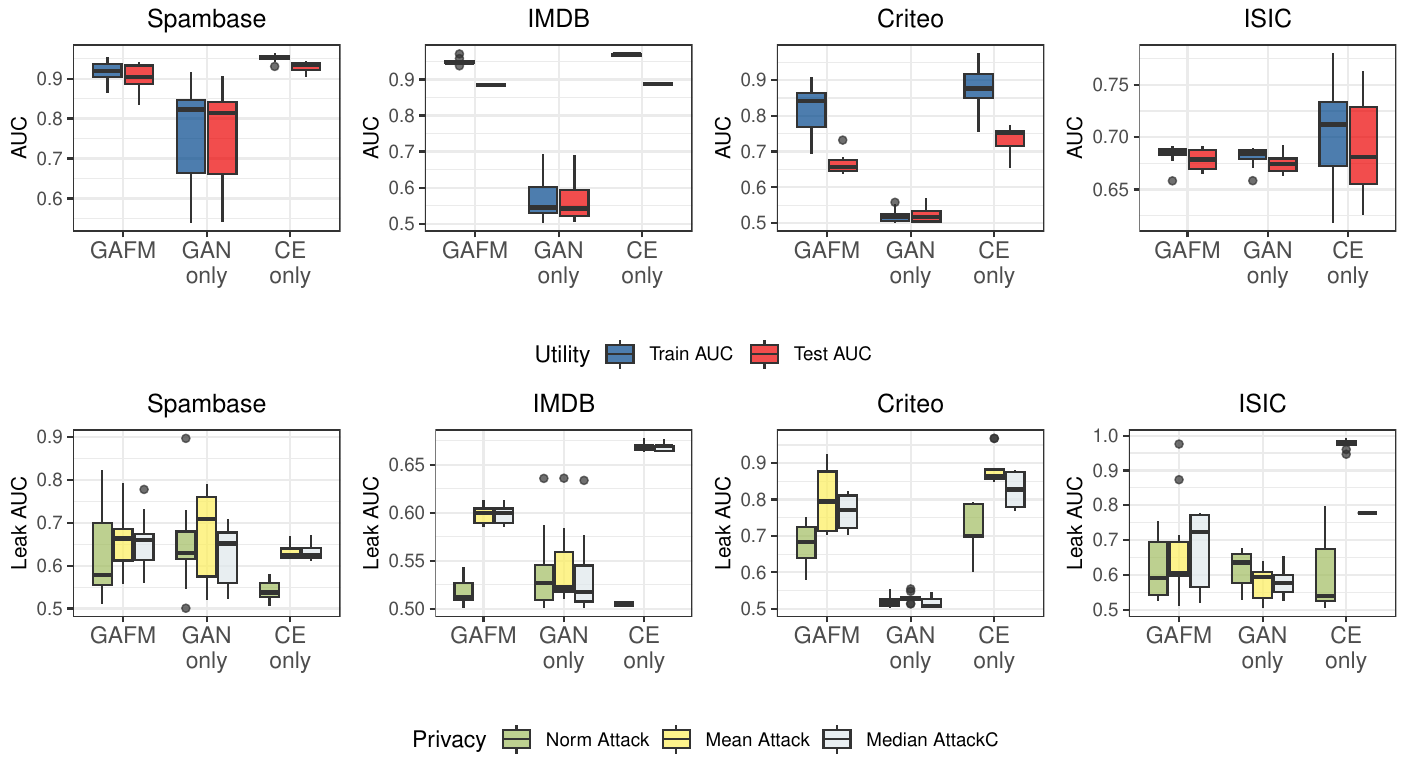}
\caption{\textbf{Comparison of GAFM with two ablated versions}: GAN-only (GAFM with only GAN loss) and CE-only (GAFM with only CE loss). Figure \ref{fig:ablation study} reveals that GAFM offers better trade-off between utility and privacy compared to the GAN-only model and the CE-only model. The GAN-only model achieves low leak AUC but has inferior classification AUC compared to GAFM. In contrast, the CE-only model fails to provide label privacy protection on most datasets. Figure \ref{fig:ablation study} demonstrates the necessity of combining both the GAN and CE components.}
\label{fig:ablation study}
\end{figure}

Figure \ref{fig:ablation study} compares GAFM with two ablated versions: GAFM with only GAN loss (GAN-only) and GAFM with only CE loss (CE-only). The results show that GAFM strikes a better balance between utility and privacy across various datasets. In contrast, the CE-only model exhibits large variations in privacy protection, while the GAN-only model consistently performs poorly in terms of model utility. Specifically, except for the Spambase dataset, the CE-only model shows extremely high leak AUCs on all other datasets, particularly on IMDB and ISIC datasets, where the mean and median attack leak AUCs reach 0.7 or even higher. While the GAN-only model achieves low leak AUC across all four datasets, its classification AUC is inferior to that of GAFM, which benefits from the guidance provided by the CE component. Particularly, on the IMDB and Criteo datasets, both GAFM and CE-only models maintain a classification AUC above 0.7, compared to around 0.5 achieved by the GAN-only model. These results highlight the necessity of combining GAN and CE components to endow GAFM with stable mitigation to LLG without compromising utility.

\section{Discussion}
\label{Discussion}

We propose the Generative Adversarial Federated Model (GAFM), a novel method for binary classification tasks in VFL. Empirical experiments on four datasets demonstrate that GAFM is a promising method for effectively mitigating LLG. Unlike Marvell and Max Norm, which use noisy intermediate gradients to improve gradient mixing between the label and non-label parties, GAFM takes a different approach. By incorporating both the GAN loss and CE loss, we observed improved gradient mixing in GAFM compared to vanilla splitNN. Our analysis further reveals that GAFM's ability to defend against LLG is attributed to the mutual gradient perturbation of the GAN loss and the CE loss. 


\textbf{Limitation and future work.} Although heuristic proof and additional experiments are provided to explain GAFM's privacy protection capabilities arise from the mutually perturbing gradients generated by the GAN and CE components, GAFM is not as rigorous as Marvell in providing an upper bound on label leakage from gradients. However, GAFM and Marvell are not mutually exclusive, and GAFM can incorporate optimized random noise provided by Marvell to enhance privacy protection. We provide an example on the Criteo dataset in Appendix \ref{The combination of GAFM and Marvell} to demonstrate that combining GAFM and Marvell offers a better mitigation of LLG compared to using GAFM or Marvell alone, albeit with a slight decrease in utility. Another consideration is whether the distinct features observed in binary classification tasks, such as notable differences between positive and negative instance gradients, persist in multi-class settings, and how to extend current methods to multi-class scenarios. This presents an exciting direction for future work.




\newpage
\bibliographystyle{plain}
\bibliography{mybib.bib}

\newpage
\appendix
\counterwithin{figure}{section}
\counterwithin{table}{section}
\counterwithin{equation}{section}
\counterwithin{algorithm}{section}

\section{GAFM Algorithm Description}
\label{GAFM Algorithm Description}

The details of GAFM are outlined in Algorithm \ref{alg:GAFM}.

\begin{algorithm}[H]
\caption{Generative Adversarial Federated Model}
\label{alg:GAFM}

\textbf{Input:}
Training data $X_1,X_2,\dots,X_P$, training labels $Y$, sample $u \sim \text{Uniform}(0, \Delta)$, $\varepsilon \sim \mathcal{N}(0, \sigma^2)$, epoch $T$, weight $\gamma$, learning rates $\alpha_d,\alpha_g,\alpha_L$, and clip value $c$.

\textbf{Output:}
Updated intermediate results $f(X)$.

\begin{algorithmic}[1]
\STATE Initialize model parameters $\theta^{0}_d$, $\theta^{0}_g$, and $\theta^{0}_p$ and calculate the initialized intermediate result $f(X)$.

\WHILE {Not converged}
     \STATE \textbf{Label party}
     \STATE \textbf{Step 1: Update the discriminator}
     \STATE The label party computes the prediction $\hat{Y} \gets G_{\theta_g}(f(X))$ and updates the discriminator by $\theta_d \gets \theta_d+\alpha_d \nabla_{\theta_d}L_{\text{GAN}}$ and clip $\theta_d \gets \text{clip}(\theta_d,-c,c)$.

     \STATE \textbf{Step 2: Update the generator}
     \STATE Then the label party updates the generator by $\theta_g \gets \theta_g-\alpha_g\nabla_{\theta_{g}}L_{\text{GAN}}$.

     \STATE \textbf{Step 3: Back-propagation}
     \STATE The label party backpropagates normalized gradients with respect to $f(X)$ to the passive parties: $grad_{f(X)} = \gamma \frac{\nabla_{f(X)}L_{\text{GAN}}}{\left\|\nabla_{f(X)}L_{\text{GAN}}\right\|_2} + \frac{\nabla_{f(X)}L_{\text{CE}}}{\left\|\nabla_{f(X)}L_{\text{CE}}\right\|_2}$.

     \STATE \textbf{Non-label parties}
     \FOR{$p=1$ to $P$}
         \STATE \textbf{Step 4: Update local models}
         \STATE Based on the locally model and $grad_{f(X)}$, passive party $p$ updates $\theta_p$.
     \ENDFOR
\ENDWHILE

\STATE \textbf{Step 5: Forward-propagation}
\STATE Update intermediate results $f(X) \gets \sum_{p=1}^P f_{\theta_p}(X_p)$.

\end{algorithmic}
\end{algorithm}

\section{Heuristic Justification for Improved Gradients Mixing}
\label{Heuristic Justification for Improved Gradients Mixing}
In this section, we provide a heuristic justification for why the GAN and CE components in GAFM tend to have opposite directions.
\begin{itemize}
    \item To simplify the notation, we assume that the intermediate result $f(X) \in \mathbb{R}^d$, where $d=1$ (the same heuristic justification can be easily applied for $d \neq 1$), and we follow the experimental setting by using the identity function $I(.)$ as the transformation function $F(.)$.
     \item We assume that $\sigma(f(X))$ (also $f(X)$) to be increase with $Y$ since it tries to match $\tilde{Y}$, which tends to increase with $Y$ itself.
\end{itemize}
To optimize the equation (\ref{eq:minmax_full_obj}), for the sample $i$ , the intermediate gradient is:
\begin{equation}
\label{grad-1}
\begin{split}
\frac{\partial L_{\mathrm{GAFM}}}{\partial f(X_i)}=\gamma \frac{\partial L_{\mathrm{GAN}}}{\partial f(X_i)}+\frac{\partial L_{\mathrm{CE}}}{\partial f(X_i)}.
\end{split}
\end{equation}
For the first term of equation (\ref{grad-1}) and assume the sample size $N$:
\begin{equation}
\label{grad-gan}
\begin{split}
\frac{\partial L_{\mathrm{GAN}}}{\partial f(X_i)}=-\frac{\partial D_{\theta_d}(G_{\theta_g}(f(X_i)))}{N \partial f(X_i)}=-\frac{\partial D_{\theta_d}(\hat{Y_i})}{N \partial \hat{Y_i}}\frac{\partial \hat{Y_i}}{\partial f(X_i)}.
\end{split}
\end{equation}
\begin{itemize}
\item  Consider the original GAN loss $L_{\mathrm{GAN}}$ and approximate it roughly with Taylor expansion and $\hat{Y} \to Y+\varepsilon$:
\begin{equation}
\label{grad-gan-2}
\begin{split}
  &L_{\mathrm{GAN}}(\theta_d,\theta_g;\varepsilon)=\mathbb E\left[D_{\theta_d}(Y+\varepsilon)\right]-\mathbb E\left[D_{\theta_d}(G_{\theta_g}(f(X_i)))\right]\\
&= \mathbb E_{Y=1}[ D_{\theta_d}(1+\varepsilon)-D_{\theta_d}(\hat Y)]+\mathbb E_{Y=0}[ D_{\theta_d}(\varepsilon)-D_{\theta_d}(\hat Y)]\\ 
&\approx \mathbb E_{Y=1}[ \nabla D_{\theta_d}(\hat Y)(1+\varepsilon - \hat Y)]+\mathbb E_{Y=0}[\nabla D_{\theta_d}(\hat Y)(\varepsilon - \hat Y)]\\
&\approx \mathbb E_{Y=1}[ \nabla D_{\theta_d}(\hat Y)(1- \hat Y)]-\mathbb E_{Y=0}[ \nabla D_{\theta_d}(\hat Y)\hat Y].
\end{split}
\end{equation}
Hence,  heuristically, $\nabla D_{\theta^d}(\hat Y_i)$ tends to be positive and large if $\hat Y_i$ is large, whereas $\nabla D_{\theta^d}(\hat Y_i)$ tends to be small and negative  if $\hat Y_i$ is small.  Since $\hat Y$ is increasing with $Y$, we tend to have (1) $\frac{\partial \hat Y}{\partial f(X)} > 0$, and (2) and $\nabla D_{\theta_d}(\hat Y_1) - \nabla D_{\theta_d}(\hat Y_0) > 0$, where $\hat Y_1$, $\hat Y_0$ represent some $\hat Y$  from class 1 and class 0 respectively. Hence, $\frac{\partial L_{\mathrm{GAN}}}{\partial f_1(X)}$ tend to be smaller than $\frac{\partial L_{\mathrm{GAN}}}{\partial f_0(X)}$ where  $f_1(X)$, $f_0(X)$ represent some $f(X)$  from class 1 and class 0. This is GAN loss gradients of GAFM we observed in figure \ref{fig:grads on IMDB}, figure \ref{app:grads on spambase} and figure \ref{app:grads on criteo}.
\item  Consider the original GAN loss $L_{GAN}$ and approximate it roughly with Taylor expansion and $\hat{Y} \to 1-(Y+\varepsilon)$, we have the similar conclusion as equation \ref{grad-gan-2}. Since  $\hat Y$ is decreasing with $Y$,  we tend to have (1) $\frac{\partial \hat Y}{\partial f(X)} < 0$, and (2) and $\nabla D_{\theta_d}(\hat Y_0) - \nabla D_{\theta_d}(\hat Y_1) > 0$. Hence, we still have $\frac{\partial L_{\mathrm{GAN}}}{\partial f_1(X)}$ tend to be smaller than $\frac{\partial L_{\mathrm{GAN}}}{\partial f_0(X)}$. 
\end{itemize}

For the second term of equation (\ref{grad-1}): let $S_i=\sigma(f(X_i))$ and $\frac{\partial S_i}{\partial f(X_i)}>0$, hence,
\begin{equation}
\label{grad-CE}
\begin{split}
\frac{\partial L_{\mathrm{CE}}}{\partial f(X_i)}=\frac{1}{N}(-\frac{\tilde{Y}}{S_i}+\frac{1-\tilde{Y}}{1-S_i})\frac{\partial S_i}{\partial f(X_i)} \propto \frac{1}{N}\frac{S_i-\tilde{Y}}{S_i(1-S_i)}.
\end{split}
\end{equation}
Without the GAN loss part, in a perfectly fitted model, we tend to have,
\begin{equation}
\label{grad-nogan}
\begin{split}
\left\{\begin{array}{c}
\mathbb E(S_i|Y_i=1)=\mathbb E(\tilde{Y_i}|Y_i=1)=0.5+\frac{\Delta}{2}\\
\mathbb E(S_i|Y_i=0)=\mathbb E(\tilde{Y_i}|Y_i=0)=0.5-\frac{\Delta}{2}
\end{array}\right. .
\end{split}
\end{equation}
    

In practice, an imperfect fit tends to have $E(\Tilde{Y_i}|Y_i=1)<E(\dot{Y_i}|Y_i=1)$ and  $E(\Tilde{Y_i}|Y_i=0)>E(\dot{Y_i}|Y_i=0)$. In addition, with the GAN loss part, the larger gradient (usually) for $Y=0$ from the GAN loss drives $\tilde{Y}_i$  to decrease more at $Y=0$ compared to that from $Y = 1$. Combining them together, heuristically, we tend to have,
\begin{equation}
\label{grad-CE-2}
\begin{split}
\frac{\partial L_{\mathrm{CE}}}{\partial f(X_i)} \propto
\left\{\begin{array}{cc}
    \frac{1}{N}\frac{S_i-\tilde{Y}}{S_i(1-S_i)}>0;&Y_i=1 \\
    \frac{1}{N}\frac{S_i-\tilde{Y}}{S_i(1-S_i)}<0;  &Y_i=0
\end{array}\right.
\end{split}
\end{equation}

This is CE loss gradients of GAFM we observed in figure \ref{app:grads on spambase} and figure \ref{fig:grads on IMDB}.

Finally, combining equations (\ref{grad-gan}) and (\ref{grad-CE-2}), when $\hat{Y} \to Y+\varepsilon$, the normalized final gradient for $L_{\mathrm{GAFM}}$ is 
\begin{equation}
\label{grad-final-1}
\begin{split}
\frac{\partial L_{\mathrm{GAFM}}}{\partial f(X_i)} = \left\{\begin{array}{cc}\gamma \frac{-\nabla D_{\theta^d}(\hat{Y_i})\frac{\partial \hat{Y_i}}{\partial f(X_i)}}{\|\nabla D_{\theta^d}(\hat{Y})\frac{\partial \hat{Y}}{\partial f(X)}\|_2}\downarrow + \frac{\frac{S_i-\tilde{Y}}{NS_i(1-S_i)}\frac{\partial S_i}{\partial f(X_i)}}{\|\frac{S-\tilde{Y}}{NS_i(1-S)}\frac{\partial S}{\partial f(X)}\|_2}\uparrow ;&Y=1\\
\gamma \frac{-\nabla D_{\theta^d}(\hat{Y_i})\frac{\partial \hat{Y_i}}{\partial f(X_i)}}{\|\nabla D_{\theta^d}(\hat{Y})\frac{\partial \hat{Y}}{\partial f(X)}\|_2}\uparrow + \frac{\frac{S_i-\tilde{Y}}{NS_i(1-S_i)}\frac{\partial S_i}{\partial f(X_i)}}{\|\frac{S-\tilde{Y}}{NS_i(1-S)}\frac{\partial S}{\partial f(X)}\|_2}\downarrow;  & Y=0
\end{array}\right. ,
\end{split}
\end{equation}
where the normalization operation preserves the order of gradients. By equation \ref{grad-final-1}, our heuristic analysis suggests that the gradients from the GAN component and the CE component tend to have opposite directions, resulting in the mutual perturbation of the gradients of both the GAN loss and the CE loss, leading to a better blending of the overall gradient.

\section{Complete Experimental Results}

\subsection{Intermediate Gradients on Additional Datasets}
\label{Exploring Intermediate Gradients on Additional Datasets}

\begin{figure}[]
\centering
\centering
\includegraphics[width=14cm,height=3.6cm]{
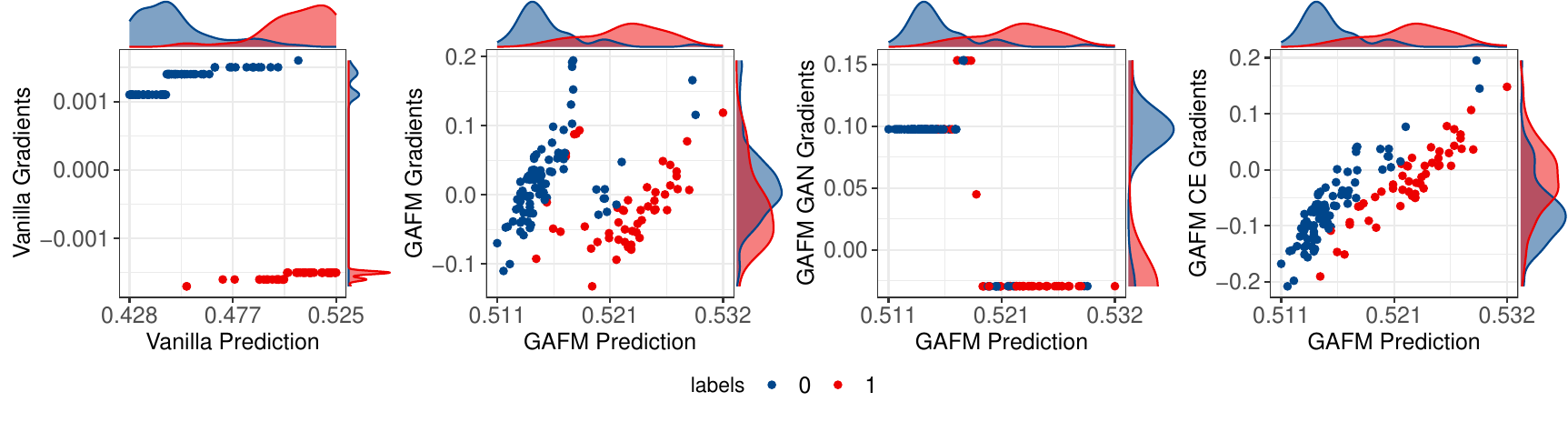}
\caption{Comparison of Prediction vs. Intermediate Gradients between vanilla SplitNN and GAFM on Spambase. The figure displays, from left to right, the intermediate gradients of vanilla SplitNN, GAFM, the GAN loss gradient of GAFM, and the CE loss gradient of GAFM. We observe that the mutual perturbation between the GAN loss gradient and the CE loss gradient generate intermediate gradients that are more mixed compared to vanilla SplitNN.}
\label{app:grads on spambase}
\end{figure}

\begin{figure}[]
\centering
\includegraphics[width=14cm,height=3.6cm]{
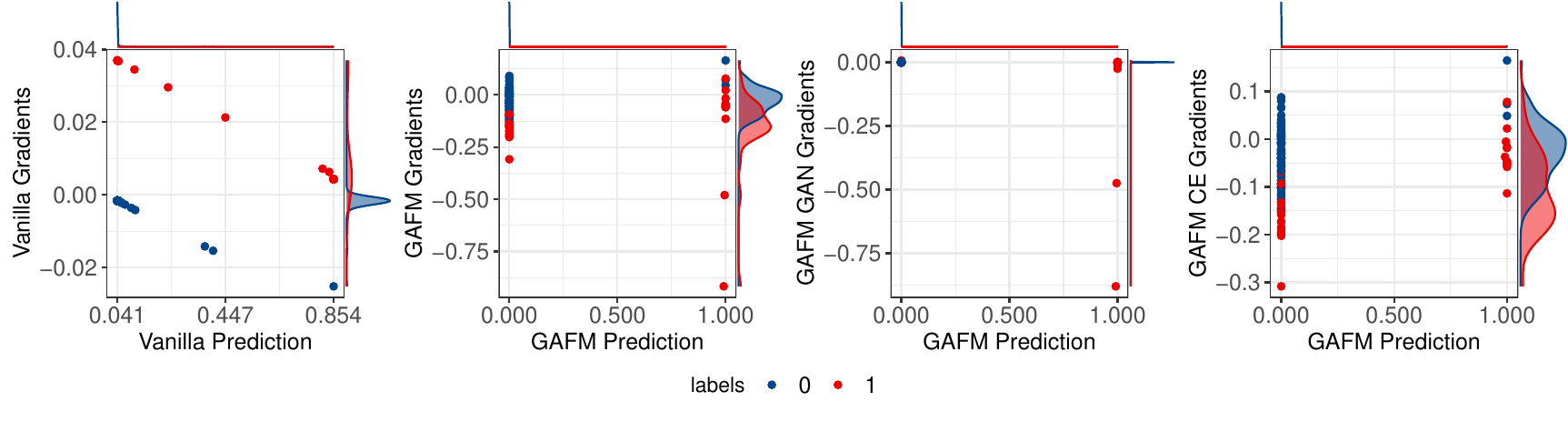}
\caption{Comparison of Prediction vs. Intermediate Gradients between vanilla SplitNN and GAFM on Criteo. The figure displays, from left to right, the intermediate gradients of vanilla SplitNN, GAFM, the GAN loss gradient of GAFM, and the CE loss gradient of GAFM. We observe that the mutual perturbation between the GAN loss gradient and the CE loss gradient generate intermediate gradients that are more mixed compared to vanilla SplitNN.}
\label{app:grads on criteo}
\end{figure}

\begin{figure}[]
\centering
\setlength{\abovecaptionskip}{-0.8cm}
\centering
\includegraphics[width=14cm,height=3.6cm]{
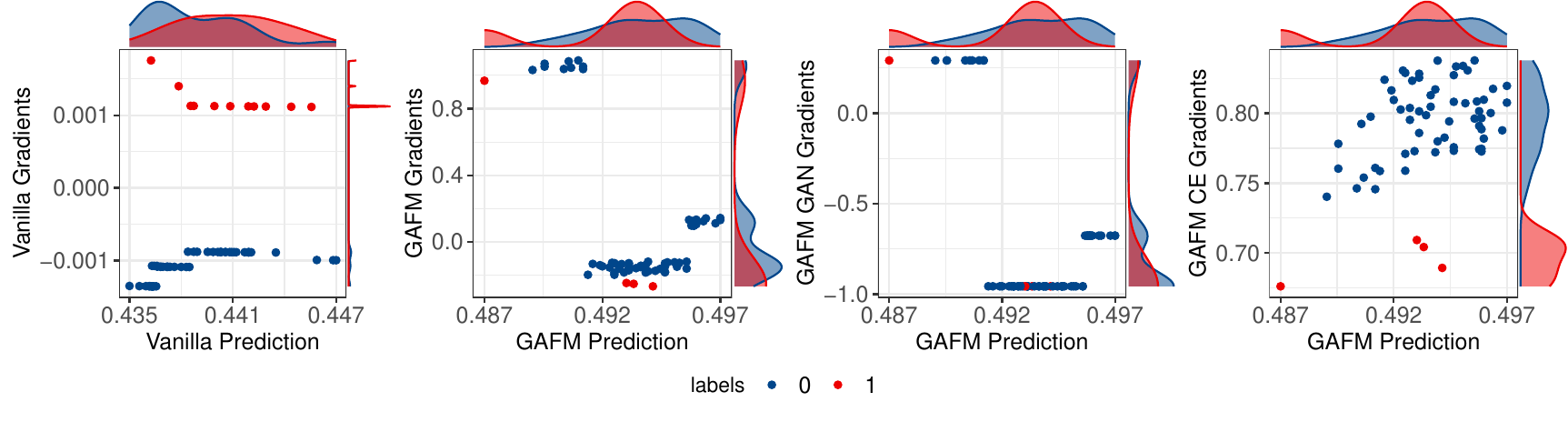}
\caption{Comparison of Prediction vs. Intermediate Gradients between vanilla SplitNN and GAFM on ISIC. The figure displays, from left to right, the intermediate gradients of vanilla SplitNN, GAFM, the GAN loss gradient of GAFM, and the CE loss gradient of GAFM. We observe that the mutual perturbation between the GAN loss gradient and the CE loss gradient generate intermediate gradients that are more mixed compared to vanilla SplitNN.}
\label{app:grads on ISIC}
\end{figure}

Figures \ref{app:grads on spambase}, \ref{app:grads on criteo}, and \ref{app:grads on ISIC} show the intermediate gradients of Spambase, Criteo, and ISIC datasets, respectively. Consistent with the results on the IMDB dataset, we observe that GAFM has more mixed gradients compared to Vanilla on these three datasets. These observation support the analysis in Appendix \ref{Heuristic Justification for Improved Gradients Mixing} that the mixed gradients of GAFM can be attributed to the mutual cancellation between GAN loss gradients and CE loss gradients.

\subsection{Evaluation in the Multi-Party Setting}
\label{Multiple Clients}

In this section, we extend the two-party split learning setting to the multi-party split learning setting, where three non-label parties and one label party collaborate to train GAFM. We evaluate two scenarios for local feature assignment: balanced and imbalanced. For Spambase and Criteo datasets, the local feature assignment ratio among the three participants is set to $client_1:client_2:client_3=1:1:1$, while for IMDB dataset, it is set to $client_1:client_2:client_3=2:2:1$. Since ISIC image dataset is not suitable for multi-party scenarios, we exclude it from our experiments. The experimental setup is consistent with that of section \ref{Experiment Setup}, with the exception of the transformation function $F(.)$. In this case, $F(.)$ is an averaging function rather than the identity function $I(.)$.

\begin{equation}
\label{averaging function}
\begin{split}
f(X)=F(f(X_1),f(X_2),f(X_3))=\frac{f(X_1)+f(X_2)+f(X_3)}{3}.
\end{split}
\end{equation}

where $f(X_i)$ is the intermediate results from the $ith$ non-label party.

\begin{table}[]
\centering
\caption{Comparison of utility and privacy on multi-party setting over 10 repetitions. In the multi-party setting, GAFM still outperforms Vanilla and Max Norm in terms of privacy preservation, and Marvell in terms of utility.}
\label{app:auc-leak AUC-tvd-mclient}
\begin{tabular}{lccccccr}
\toprule
dataset&Method&\makecell{Training\\ AUC}&\makecell{Testing\\ AUC} & \makecell{Norm\\ Attack} & \makecell{Mean\\ Attack} & \makecell{Median\\ Attack}\\
\midrule
\multirow{4}*{Spambase}&GAFM&  0.87$\pm$0.15&  0.86$\pm$0.15&0.60$\pm$0.07& {0.69$\pm$0.06}& {0.68$\pm$0.07}\\
&Marvell & 0.70$\pm$0.05&0.68$\pm$0.06&{0.53$\pm$0.03}& {0.70$\pm$0.02}& {0.70$\pm$0.02}\\ 
&Max Norm&{0.95$\pm$0.00}&{0.95$\pm$0.00}& 0.83$\pm$0.11& 0.99$\pm$0.02& 0.99$\pm$0.04\\
&Vanilla&{0.95$\pm$0.00}&{0.95$\pm$0.00}& 0.84$\pm$0.13& 1.00$\pm$0.00& 1.00$\pm$0.00\\
\midrule
\multirow{4}*{IMDB}&GAFM& 0.97$\pm$0.01& 0.82$\pm$0.00& {0.54$\pm$0.02}& {0.65$\pm$0.04}&{0.54$\pm$0.02}\\
&Marvell&0.82$\pm$0.01& 0.78$\pm$0.01 & {0.51$\pm$0.01}& {0.71$\pm$0.02}& {0.71$\pm$0.02}\\ 
&Max Norm&{0.97$\pm$0.00}& {0.88$\pm$0.00}& 0.54$\pm$0.03& 0.99$\pm$0.01& 1.00$\pm$0.00\\
&Vanilla&{0.97$\pm$0.00}& {0.88$\pm$0.00}& 0.54$\pm$0.02& 0.99$\pm$0.01& 1.00$\pm$0.00\\
\midrule
\multirow{4}*{Criteo}&GAFM&  0.80$\pm$0.01&  0.65$\pm$0.02&0.63$\pm$0.01&  {0.86$\pm$0.01}& {0.80$\pm$0.00}\\
&Marvell & 0.83$\pm$0.03&0.70$\pm$0.04&{0.75$\pm$0.08}& {0.85$\pm$0.09}& {0.78$\pm$0.04}\\ 
&Max Norm&{0.83$\pm$0.02}&{0.71$\pm$0.02}& 0.91$\pm$0.03& 0.95$\pm$0.09& 0.82$\pm$0.00\\
&Vanilla&{0.83$\pm$0.02}&{0.71$\pm$0.02}& 0.91$\pm$0.03& 0.95$\pm$0.09& 0.82$\pm$0.00\\
\bottomrule
\end{tabular}
\end{table}
The leak AUC in Tabel \ref{app:auc-leak AUC-tvd-mclient} measures not only the overall leakage, but also the leakage of each non-label party, as the $\Tilde{y}_1$, $\Tilde{y}_2$, and $\Tilde{y}_3$ are equally weighted at the cut layer. We observe that the utility of both GAFM and Marvell slightly decreases in the multi-party setting compared to the two-party setting, possibly due to the increased difficulty of jointly training with multiple participants. However, in the multi-client setting, GAFM still achieves a better trade-off between utility and privacy compared to all baselines.

\section{Discussion on hyperparameters}
\label{Choice of model parameters}
In this section, we present the methods for selecting the key hyperparameters $\sigma$, $\gamma$, and $\Delta$ in GAFM. Table \ref{table:param} presents the hyperparameters of GAFM that are utilized in our experiments on four datasets.

\begin{table}[]
\centering
\caption{Hyperparameters of GAFM on four datasets.The following section discusses the selection of three crucial parameters in GAFM: $\sigma$ (section \ref{Discussion on sigma}), $\gamma$ (section \ref{Discussion on gamma}), and $\Delta$ (section \ref{Discussion on delta}), respectively.}
\label{table:param}
\begin{tabular}{cccccc}
\hline
Dataset & $\sigma$&$\Delta$ & $ \gamma$ \\ \hline
Spambase& 0.01&0.05&1 \\
IMDB&   0.01&0.1&1 \\ 
Criteo&  0.01&0.5&1\\
ISIC&  0.01&0.05&20
\\ \hline
\end{tabular}
\end{table}

\subsection{Discussion on $\sigma$}
\label{Discussion on sigma}

We conducted a comparison of GAFM with varying values of $\sigma$ over 10 repetitions, as shown in Table \ref{app:auc-leak AUC-tvd-sigma}. Our results indicate that a moderately large $\sigma$ does not significantly compromise prediction accuracy and can still offer robust protection against label stealing attacks. Thus, we have chosen to fix $\sigma$ at 0.01 for all datasets.
\begin{table}[]
\centering
\caption{Average AUC and Leak AUC of GAFM with different $\sigma$. We observe that GAFM is insensitive to  $\sigma$ in terms of both utility and privacy.}
\label{app:auc-leak AUC-tvd-sigma}
\begin{tabular}{lccccccccr}
\toprule
dataset&$\sigma$&\makecell{Training\\ AUC} &\makecell{Test\\ AUC}& \makecell{Norm\\ Attack} & \makecell{Mean\\ Attack} & \makecell{Median\\ Attack}\\
\midrule
\multirow{3}*{Spambase}&0.01&  0.94$\pm$0.01&  0.93$\pm$0.02&{0.56$\pm$0.04}&{0.67$\pm$0.05}&{0.66$\pm$0.05}\\
&0.25& 0.94$\pm$0.01&0.93$\pm$0.01&{0.60$\pm$0.06}& {0.67$\pm$0.03}& {0.67$\pm$0.03}\\ 
&1&{0.94$\pm$0.01}&{0.94$\pm$0.01}& 0.57$\pm$0.05& 0.68$\pm$0.04& 0.68$\pm$0.04\\
\midrule
\multirow{3}*{IMDB}&0.01& 0.95$\pm$0.01& 0.88$\pm$0.00&{0.52$\pm$0.01}&{0.60$\pm$0.01}& {0.60$\pm$0.01}\\
&0.25&0.92$\pm$0.14& 0.85$\pm$0.11& {0.52$\pm$0.02}& {0.60$\pm$0.03}& {0.61$\pm$0.03}\\ 
&1&{0.87$\pm$0.18}& 0.81$\pm$0.15& {0.54$\pm$0.03}& 0.60$\pm$0.06& 0.61$\pm$0.05\\
\midrule
\multirow{3}*{Criteo}&0.01& 0.82$\pm$0.07&0.67$\pm$0.03&{0.68$\pm$0.06}&{0.80$\pm$0.09}& {0.77$\pm$0.05}\\
&0.25&0.80$\pm$0.00& 0.69$\pm$0.03& {0.74$\pm$0.07}& {0.82$\pm$0.03}& {0.82$\pm$0.00}\\ 
&1&{0.81$\pm$0.02}& 0.66$\pm$0.03& {0.71$\pm$0.08}& 0.82$\pm$0.02& 0.82$\pm$0.00\\
\midrule
\multirow{3}*{ISIC}&0.01&0.68$\pm$0.01&0.68$\pm$0.01&0.62$\pm$0.09&{0.66$\pm$0.15}&{0.68$\pm$0.11}\\
&0.25&0.68$\pm$0.01& 0.67$\pm$0.01& {0.63$\pm$0.06}& {0.69$\pm$0.13}& {0.69$\pm$0.09}\\ 
&1&{0.68$\pm$0.01}& 0.67$\pm$0.01& {0.63$\pm$0.06}& 0.69$\pm$0.13& 0.69$\pm$0.09\\
\bottomrule
\end{tabular}
\end{table}

\subsection{Discussion on $\gamma$}
\label{Discussion on gamma}

We use the loss weight parameter $\gamma$ to balance the GAN and CE loss gradients in GAFM, allowing them to effectively perturb each other. Table \ref{table:grads_mean} shows the average gradients of the GAN and CE losses for the GAFM model on four datasets, with $\gamma=1$ and other hyperparameters consistent with the section \ref{Experiment Setup}. The results indicate that the mean gradients of the GAN and CE losses are similar in magnitude for the Spambase, IMDB, and Criteo datasets, but the CE loss gradient is over 10 times larger than the GAN loss gradient for the ISIC dataset. This explains why Spambase, IMDB, and Criteo can achieve satisfactory utility-privacy trade-offs with $\gamma=1$, while ISIC requires a higher value of $\gamma$.

\begin{table}[]
\centering
\caption{Results of the average GAN loss gradient and average CE loss gradient on the four datasets. It is observed that only for the ISIC dataset, the CE loss gradient mean and GAN loss gradient mean are not on the same scale. To better achieve gradient perturbation, an increase in $\gamma$  for ISIC is necessary.}
\label{table:grads_mean}
\begin{tabular}{lccccccccr}
\toprule
dataset&($\Delta, \gamma,\sigma$)&\makecell{GAN Gradient\\ Avg.} &\makecell{CE Gradient\\ Avg.}& \makecell{Comparable \\Magnitude}\\
\midrule
Spambase&(0.05,1,0.01)& 0.054&-0.049&$\checkmark$\\
IMDB&(0.05,1,0.01)&0.006&-0.006&$\checkmark$\\ 
Criteo&(0.05,1,0.01)&-0.011&0.048&$\checkmark$\\
ISIC&(0.05,1,0.01)&0.058& 0.789&$\times$\\

\bottomrule
\end{tabular}
\end{table}

We further discuss how to select the appropriate value of $\gamma$ for the ISIC dataset. Based on the results shown in Table \ref{table:grads_mean}, the value of $\gamma$ should ideally be in the range of 10 to 20. We set the parameter $\Delta$ to a minimum value of 0.05 to control its impact on privacy and utility while keeping all other parameters and experimental settings unchanged. Table \ref{table:ISIC_gamma_subset} reports the experimental results of GAFM's privacy and utility under different values of $\gamma$. 
GAFM achieves comparable average training AUC and test AUC on the ISIC 10\% subset across the range of $\gamma$ values. However, at $\gamma=20$, GAFM achieves the minimum average leak AUC under three different attacks. Therefore, we conclude that the optimal parameter for the ISIC subset is $\gamma=20$.

\begin{table}[]
\centering
\caption{The impact of different $\gamma$ values on the utility and privacy of GAFM on \textbf{the 10\% subset ISIC data}. Table \ref{table:ISIC_gamma_subset} reports the experimental results of GAFM with various $\gamma$ values, while maintaining all other experimental settings consistent. Moreover, we set the parameter $\Delta$ to a minimum value of 0.05 to control its effect on utility and privacy. Based on the overall assessment of utility and privacy, we conclude that the optimal value for $\gamma$ is 20.}
\label{table:ISIC_gamma_subset}
\begin{tabular}{lccccccccr}
\toprule
$\gamma$&($\Delta, \sigma$)&\makecell{Training\\ AUC} &\makecell{Test\\ AUC}& \makecell{Norm\\ Attack} & \makecell{Mean\\ Attack} & \makecell{Median\\ Attack}\\
\midrule
1& (0.05,0.01)&0.65$\pm$0.06&0.65$\pm$0.06&0.77$\pm$0.04&{0.99$\pm$0.01}&{0.78$\pm$0.00}\\ 
10&(0.05,0.01)&0.67$\pm$0.03&0.66$\pm$0.05&0.77$\pm$0.04&{0.98$\pm$0.02}&{0.78$\pm$0.00}\\
15&(0.05,0.01)&0.67$\pm$0.03&0.66$\pm$0.05&0.76$\pm$0.04&{0.96$\pm$0.04}&{0.77$\pm$0.00}\\
20&(0.05,0.01)&0.67$\pm$0.03&0.66$\pm$0.05&0.76$\pm$0.04&{0.93$\pm$0.05}&{0.76$\pm$0.00}\\
\bottomrule
\end{tabular}
\end{table}

To validate the effectiveness of the selected $\gamma$ on the subset, we also report the experimental results on the full ISIC dataset with different values of $\gamma$ in Table \ref{table:ISIC_gamma_full}. We find that $\gamma=20$ is also an appropriate value for the full dataset. The consistency of the selected $\gamma$ for both the subset and the full dataset demonstrates the validity of determining $\gamma$ based on sampling.

\begin{table}[]
\centering
\caption{The impact of different $\gamma$ values on the utility and privacy of GAFM on \textbf{the full  ISIC data}. Table \ref{table:ISIC_gamma_subset} reports the experimental results of GAFM with various $\gamma$ values, while maintaining all other experimental settings consistent. Moreover, we set the parameter $\Delta$ to a minimum value of 0.05 to control its effect on utility and privacy. Based on the overall assessment of utility and privacy, we conclude that the optimal value for $\gamma$ is 20.}
\label{table:ISIC_gamma_full}
\begin{tabular}{lccccccccr}
\toprule
$\gamma$&($\Delta, \sigma$)&\makecell{Training\\ AUC} &\makecell{Test\\ AUC}& \makecell{Norm\\ Attack} & \makecell{Mean\\ Attack} & \makecell{Median\\ Attack}\\
\midrule
1& (0.05,0.01)&0.67$\pm$0.02&0.66$\pm$0.03&0.98$\pm$0.01&{0.97$\pm$0.02}&{0.78$\pm$0.00}\\ 
10&(0.05,0.01)&0.67$\pm$0.02&0.68$\pm$0.01&0.77$\pm$0.19&{0.80$\pm$0.12}&{0.74$\pm$0.04}\\
15&(0.05,0.01)&0.67$\pm$0.04&0.66$\pm$0.05&0.61$\pm$0.07&{0.66$\pm$0.15}&{0.68$\pm$0.09}\\
20&(0.05,0.01)&0.68$\pm$0.01&0.68$\pm$0.01&0.62$\pm$0.09&{0.66$\pm$0.15}&{0.68$\pm$0.11}\\
\bottomrule
\end{tabular}
\end{table}

\subsection{Discussion on $\Delta$}
\label{Discussion on delta}

We now turn to the topic of determining the randomized response-related hyperparameter $\Delta$ in GAFM. In section \ref{Experiment Setup}, we chose $\Delta$ based on the measure $Ratio=\frac{leak AUC}{train AUC}$ and the minimum average ratio criterion. Specifically, we selected $\Delta$ that yields the lowest average ratio across all three attacks.

We follow the procedure outlined below. After fixing the other parameters, we only randomly sample a small subset (10\%) of the dataset without replacement to ensure privacy, which is shared between the label and non-label parties. GAFM is trained on this small subset with different values of $\Delta$, and we repeat this process 5 times. We then compute the ratio of norm attack, mean attack, and median attack for each value of $\Delta$ and average these ratios across the repetitions. The results are presented in Table \ref{app: Delta}. Based on these ratios, we select the optimal $\Delta$ for each dataset subset. Specifically, the optimal $\Delta$ for the Spambase subset is 0.05, for the IMDB subset is 0.1, for the Criteo subset is 0.5, and for the ISIC subset is 0.05.

\begin{figure}[]
\centering
\setlength{\abovecaptionskip}{-0.2cm}
\centering
\includegraphics[width=14cm,height=7cm]{
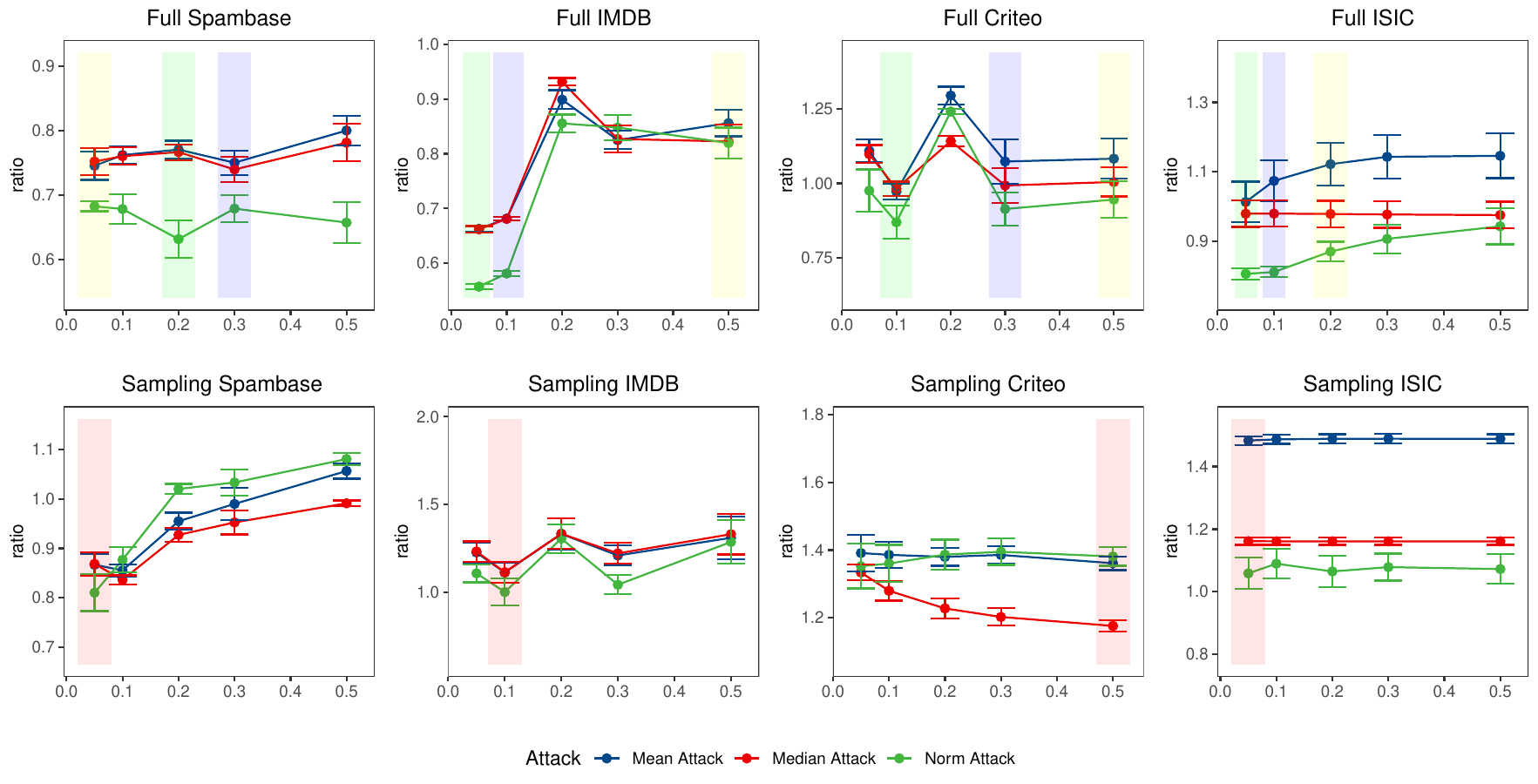}
\caption{The ratio distribution on the full dataset and small subset. In Figure \ref{fig: Delta}, the green (smallest), blue (second smallest), and orange (third smallest) blocks correspond to the three smallest ratio values on the full dataset for different $\Delta$, while the red block represents $\Delta$ with the smallest average ratio on the small subset. Notably, we observe that $\Delta$ with the smallest average ratio on the small subset always belongs to the three groups of $\Delta$ with the smallest average ratio on the full dataset.}
\label{fig: Delta}
\end{figure}

To better demonstrate the effectiveness of the ratio-based strategy, we compare the detailed ratios on both the full dataset and the small subset in Table \ref{app: Delta}. Our observation shows that the optimal $\Delta$ selected based on the ratio results from the small subset is consistent with that selected based on the full dataset experiment results for the ISIC dataset. For Spambase, IMDB, and Criteo datasets, the optimal $\Delta$ selected based on the ratio results from the small subset is very close to that selected based on the full dataset in terms of the average ratio value. Figure \ref{fig: Delta} further visualizes this degree of closeness, where the three colored blocks on the full dataset represent the three sets of parameters with the smallest ratio. We find that the optimal parameter (red block) selected on the small subset is always included in the three sets of parameters with the smallest ratio on the full dataset.

\begin{table}[]
\centering
\caption{Average ratio results on the full dataset and small subset. For the ISIC dataset, the optimal $\Delta$ selected is consistent between the full dataset and 10\% subset. Although for the Spambase, IMDB, and Criteo datasets, the optimal $\Delta$ selected differs between the full dataset and small subset, the selected ratio values are very close.}
\label{app: Delta}
\begin{tabular}{lccccccccr}
\toprule
dataset&sampling ratio &$\Delta=0.05$&$\Delta=0.1$&$\Delta=0.2$&$\Delta=0.3$& $\Delta=0.5$\\
\midrule
\multirow{2}*{Spambase}&1&{0.727}& 0.733&  \textbf{0.722}&{0.723}&0.746\\
&0.1&\textbf{0.849}& 0.856&  0.968&0.992&1.043\\
\midrule
\multirow{2}*{IMDB}&1&\textbf{0.627}&{0.648}& 0.895&0.833&0.833\\
&0.1&1.189& \textbf{1.076}&1.323&1.158&1.309\\
\midrule
\multirow{2}*{Criteo}&1&{1.061}& \textbf{0.941}&  {1.226}&{0.993}&1.011\\
&0.1&{1.359}&1.342&1.331&1.328&\textbf{1.306}\\
\midrule
\multirow{2}*{ISIC}&1&\textbf{0.933}& 0.955&  {0.990}&{1.009}&1.022\\
&0.1&\textbf{1.234}& 1.246&  1.238&1.243&1.241\\
\bottomrule
\end{tabular}
\end{table}

\section{Data setup and Experiment Details}
\label{Data setup and Experiment Details}

In this section, we provide a detailed overview of our experimental setup. We first describe the pre-processing steps for the four public datasets in \ref{dataset Processing}. Next, we provide a description of the model architecture used for each dataset in \ref{Model Architecture Details}. Finally, we present the training hyperparameters for each dataset and model combination in \ref{Model Training Details}.

\subsection{Dataset Processing}
\label{dataset Processing}

\textbf{Spambase} Spambase data are classified as whether they are spam or not. The Spambase dataset contains 4061 instances, each with 55 continuous real attributes and 2 continuous integer attributes. To preprocess the data, we replace all NA values with 0 and normalize the features to reduce the impact of magnitude differences. During each repetition, we split the dataset into a 70\%-30\% train-test split at different random seed settings.

\textbf{IMDB} The IMDB data has been preprocessed and words are encoded as a sequence of word indexes in the form of integers. We select the top 500 words and encode indexes with one-hot. The 50000 reviews are split into 25000 for training and 25000 for testing at different random seeds. And the final training and test data are both $25000 \times 500$ matrices filled with 0s and 1s.

\textbf{Criteo} The Criteo dataset consists of a portion of Criteo's traffic over a period of 7 days. Each row corresponds to a display ad served by Criteo. The dataset comprises 13 integer features and 26 categorical features, all of which have been hashed onto 32 bits. We followed the winner of the Criteo Competition's recommendations for data preprocessing\footnote{\url{https://www.csie.ntu.edu.tw/~r01922136/kaggle-2014-criteo.pdf}} and the code\footnote{\url{https://github.com/rixwew/pytorch-fm/blob/master/torchfm/dataset/criteo.py}} to process the entire dataset, rather than just a 10\% subset like the Marvell paper. Our preprocessing steps included removing infrequent features that appeared in less than 10 instances, transforming numerical features (I1-I13) into categorical features, and treating them as a single feature. Additionally, we discretized numerical values using $log2$ transformation. For each iteration, we split the dataset into an 80\%-20\% train-test split using different random seed settings.

\textbf{ISIC} The official SIIM-ISIC Melanoma Classification dataset contains a total of 33126 skin lesion images, with less than 2\% positive examples. To preprocess the ISIC data, we followed the approach outlined in this code\footnote{\url{https://github.com/OscarcarLi/label-protection/blob/main/preprocess_ISIC.ipynb}}. During each iteration, we revise the image size $64 \times 64 \times 1$ and randomly split the dataset into an 80\%-20\% train-test split, using different random seed settings for each split.

\subsection{Model Architecture Details}
\label{Model Architecture Details}

To maintain consistency across all datasets, we utilized a 2-layer DNN as the generator and discriminator for our GAFM model. The intermediate results $f(X)$ were already transformed into linear embeddings, making them easily handled by the 2-layer DNN. For the generator and discriminator, we used LeakyReLU as the activation function for all layers except for the generator's output layer, which used the Sigmoid activation function. In the following sections, we will discuss the specific local model architectures used for feature extraction on each dataset.

\textbf{Spambase} For the Spambase dataset, the local model architecture consists of two linear layers with a LeakyReLU activation function. The first layer transforms the input size to a hidden representation of $batchsize \times 16$, and the second layer maps the hidden representation to a one-dimensional output using a sigmoid function.

\textbf{IMDB} The local model architecture for the IMDB datase is a fully connected model with 3 linear layers. The model consists of two hidden layers with 256 and 128 units, respectively, followed by ReLU activation functions and a dropout rate of 0.5. The output layer has a single unit after a sigmoid function.

\textbf{Criteo} In the context of online advertising data from Criteo, we utilize a widely-used deep learning model architecture called Wide and Deep model \cite{cheng2016wide}. To implement  Wide and Deep model, we refer to the code\footnote{\url{https://github.com/BrandonCXY/Pytorch_RecommenderSystem/blob/master/DL\%20Models\%20Implementation\%20in\%20Recommender\%20Sys\%20with\%20Pytorch.py}}. The architecture of  Wide and Deep model consists of three main components: FeaturesLinear, FeaturesEmbedding, and MultiLayerPerceptron. The FeaturesLinear module is a linear model that uses embedding layers and a bias term to model feature interactions. The embeddings have a desired output dimension of 1. And the FeaturesEmbedding and MultiLayerPerceptron components take embedding layers as input features and uses a series of fully connected layers to learn complex feature interactions. In this implementation, the embedding dimension is set to 16 and the MLP dimension is set to $16 \times 16$. These components are combined in the local model, which adds a linear output layer with an output dimension of 1.

\textbf{ISIC} The local model for ISIC consists of 3 convolutional layers with 6, 16, and 32 output channels respectively. Each convolutional layer uses a $3\times3$ filter with a stride of $1\times1$. Following each convolutional layer is a ReLU activation function, and the output of each activation is max-pooled with a $2\times2$ window and stride size of $2\times2$. The output of the third convolutional layer is then flattened into size $32\times6\times6$ and passed into 3 fully connected layers, with 120, 84, and 40 units respectively.

\subsection{Model Training Details}
\label{Model Training Details}

We set the transformation function $F(.)$ to the identity function $I(.)$, clip the value at $c=0.1$, use random seeds from 0 to 9. We employ PyTorch with CPU for Spambase and IMDB and employ PyTorch with CUDA acceleration to conduct all experiments for Criteo and ISIC. The Spambase dataset takes approximately 10 minutes per repetition, while the IMDB dataset requires around 0.5 hours. For the Criteo dataset, each repetition takes about 2.5 hours, and for the ISIC dataset, it takes about 40 minutes per repetition.

\textbf{Spambase, IMDB} We use the Adam optimizer with a batch size of 1028 and a learning rate of 1e-4 throughout the entire training of 300 epochs.

\textbf{Criteo} We use the Adam optimizer with a batch size of 256 and a learning rate of 1e-4 throughout the entire training of 100 epochs.

\textbf{ISIC} We use the Adam optimizer with a batch size of 256 and a learning rate of 1e-6 throughout the entire training of 250 epochs.

\section{The Combination of GAFM and Marvell}
\label{The combination of GAFM and Marvell}

In this section, we use the Criteo dataset as an example to demonstrate that combining GAFM and Marvell can enhance the privacy protection of GAFM. Table \ref{tab:G-M} shows the utility and privacy of GAFM, Marvell, and their combination (referred to as the G-M model) on the Criteo dataset, with the same experiment setup as the section \ref{Experiment Setup}. We observe a significant decrease in the leak AUC of the G-M model compared to Marvell and GAFM, demonstrating its improved ability to mitigate LLG.

\begin{table}[]
\centering
\caption{An example of the combination of GAFM and Marvell on Criteo. Table \ref{tab:G-M} demonstrates that the G-M model outperforms GAFM and Marvell in terms of lower leak AUC, indicating a better ability to protect label privacy, despite some utility degradation.}
\label{tab:G-M}
\begin{tabular}{ccccccc}
\hline
\multirow{2}{*}{Method} & \multicolumn{3}{c}{Utility} & \multicolumn{3}{c}{Privacy}                \\ \cmidrule(lr){2-4} \cmidrule(lr){5-7}  & \makecell{Avg.\\ AUC} &\makecell{Worst\\ AUC}& \makecell{Best\\AUC} & \makecell{Norm\\ Attack} & \makecell{Mean\\ Attack} & \makecell{Median\\ Attack}\\ 
\midrule
GAFM& 0.67&0.64 &0.73&{0.68$\pm$0.06}&{0.80$\pm$0.09}& {0.77$\pm$0.05}\\
Marvell&0.70&0.65&0.76& {0.76$\pm$0.08}& {0.86$\pm$0.08}& {0.78$\pm$0.04}\\
G-M&0.64&0.62&0.65& {0.60$\pm$0.02}& {0.66$\pm$0.01}& {0.65$\pm$0.01}\\
\bottomrule
\end{tabular}
\end{table}

\end{document}